\documentclass[journal]{IEEEtran}
\ifCLASSINFOpdf
  \usepackage[pdftex]{graphicx}
  \DeclareGraphicsExtensions{.pdf,.jpeg,.png}
\else
\fi
\usepackage{url}
\usepackage{graphicx}
\usepackage{amsmath}
\usepackage[flushleft]{threeparttable}
\usepackage{tablefootnote}
\usepackage{booktabs}% http://ctan.org/pkg/booktabs
\usepackage[export]{adjustbox}
\usepackage{array}
\newcolumntype{P}[1]{>{\centering\arraybackslash}p{#1}}
\newcolumntype{M}[1]{>{\centering\arraybackslash}m{#1}}
\usepackage{dblfloatfix}

\usepackage{algorithmic}
\usepackage{algorithm}

\usepackage{subfig}
\begin{document}
\title{RaspiReader: An Open Source Fingerprint Reader Facilitating Spoof Detection}

\author{Joshua~J.~Engelsma,
        Kai~Cao,
        and~Anil~K.~Jain,~\IEEEmembership{Life~Fellow,~IEEE}% <-this % stops a space
\thanks{J. J. Engelsma, K. Cao and A. K. Jain are with the Department of Computer Science and Engineering, Michigan State University, East Lansing, MI, 48824 \protect\\E-mail: \{engelsm7, kaicao, jain\}@cse.msu.edu}% <-this % stops a space}% <-this % stops a space
}

\maketitle

\begin{abstract}
We present the design and prototype of an open source, optical fingerprint reader, called RaspiReader, using ubiquitous components. RaspiReader, a low-cost and easy to assemble reader, provides the fingerprint research community a seamless and simple method for gaining more control over the sensing component of fingerprint recognition systems. In particular, we posit that this versatile fingerprint reader will encourage researchers to explore novel spoof detection methods that integrate both hardware and software. RaspiReader's hardware is customized with two cameras for fingerprint acquisition with one camera providing high contrast, frustrated total internal reflection (FTIR) images, and the other camera outputting direct images. Using both of these image streams, we extract complementary information which, when fused together, results in highly discriminative features for fingerprint spoof (presentation attack) detection. Our experimental results demonstrate a marked improvement over previous spoof detection methods which rely only on FTIR images provided by COTS optical readers. Finally, fingerprint matching experiments between images acquired from the FTIR output of the RaspiReader and images acquired from a COTS fingerprint reader verify the interoperability of the RaspiReader with existing COTS optical readers.

\end{abstract}


% \fi
\IEEEpeerreviewmaketitle

\section{Introduction}
\IEEEPARstart{O}{ne} of the major challenges facing biometric technology is the growing threat of presentation attacks, commonly referred to as spoofing\footnote{In this paper, we use the terms presentation attack and spoof interchangeably.} \cite{odin}. In this paper, we are interested in fingerprint presentation attacks in which a {\it hacker} intentionally assumes the identity of unsuspecting individuals, called {\it victims} here, through stealing their fingerprints, fabricating spoofs with the stolen fingerprints, and maliciously attacking fingerprint recognition systems with the spoofs into identifying the hacker as the victim\footnote{Presentation attacks can also occur when (i) two individuals are in collusion or (ii) an individual obfuscates his or her own fingerprints to avoid recognition \cite{spoofs_survey}.} \cite{spoofs_survey, handbook, altered_yoon, gummy}. 

The need to prevent presentation attacks is paramount due to the monumental costs and loss of user privacy associated with spoofed systems. Consider for example India's Aadhaar program which (i) provides benefits and services to an ever growing population of over 1.17 billion residents through fingerprint and/or iris recognition \cite{india1, india2} and (ii) facilitates electronic financial transactions through the United Payments Interface (UPI) \cite{india3}.  Failure to detect presentation attacks on the Aadhaar system could cause the disruption of a commerce system affecting untold numbers of people. Also consider the United States Office of Biometric Identity Management (US OBIM) which is responsible for supporting the Department of Homeland Security (DHS) with biometric identification services specifically aimed at preventing people who pose a risk to the United States from entering the country \cite{obim}. Failure to detect spoofs on systems deployed by OBIM could result in a deadly terrorist attack\footnote{Border control fingerprint recognition systems have already been shown to be vulnerable to presentation attacks as far back as 2012 when a journalist successfully spoofed the fingerprint recognition system at the Hong Kong-China border \cite{hongkongchina}.}. Finally, almost all of us are actively carrying fingerprint recognition systems embedded within our personal smart devices. Failure to detect spoof attacks on smartphones \cite{caophone} could compromise emails, banking information, social media content, personal photos and a plethora of other confidential information.

\begin{figure}[t]
\begin{center}
\includegraphics[scale=0.2]{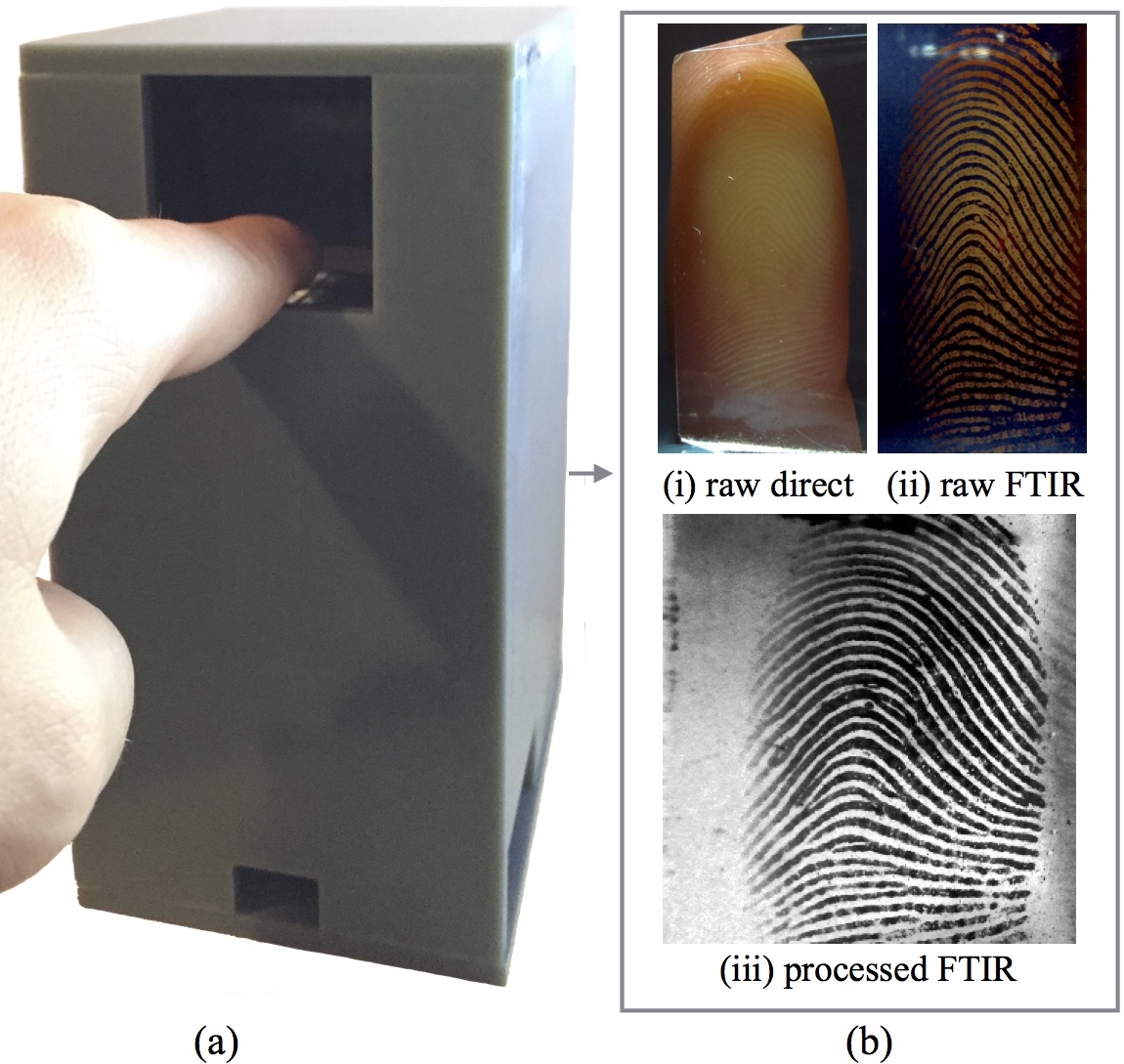}
\caption{Prototype of RaspiReader: two images (b, (i)) and (b, (ii)) of the input fingerprint (a) are captured. The raw direct image (b, (i)) and the raw, high contrast FTIR image (b, (ii)) both contain useful information for spoof detection. Following the use of (b, (ii)) for spoof detection, image calibration and processing are performed on the raw FTIR image to output a high quality, 500 ppi fingerprint for matching (b, (iii)).}
\vspace{-1.5em}
\end{center}
\label{intro_fig}
\end{figure} 

\begin{figure*}[t]
\begin{center}
\includegraphics[scale=0.25]{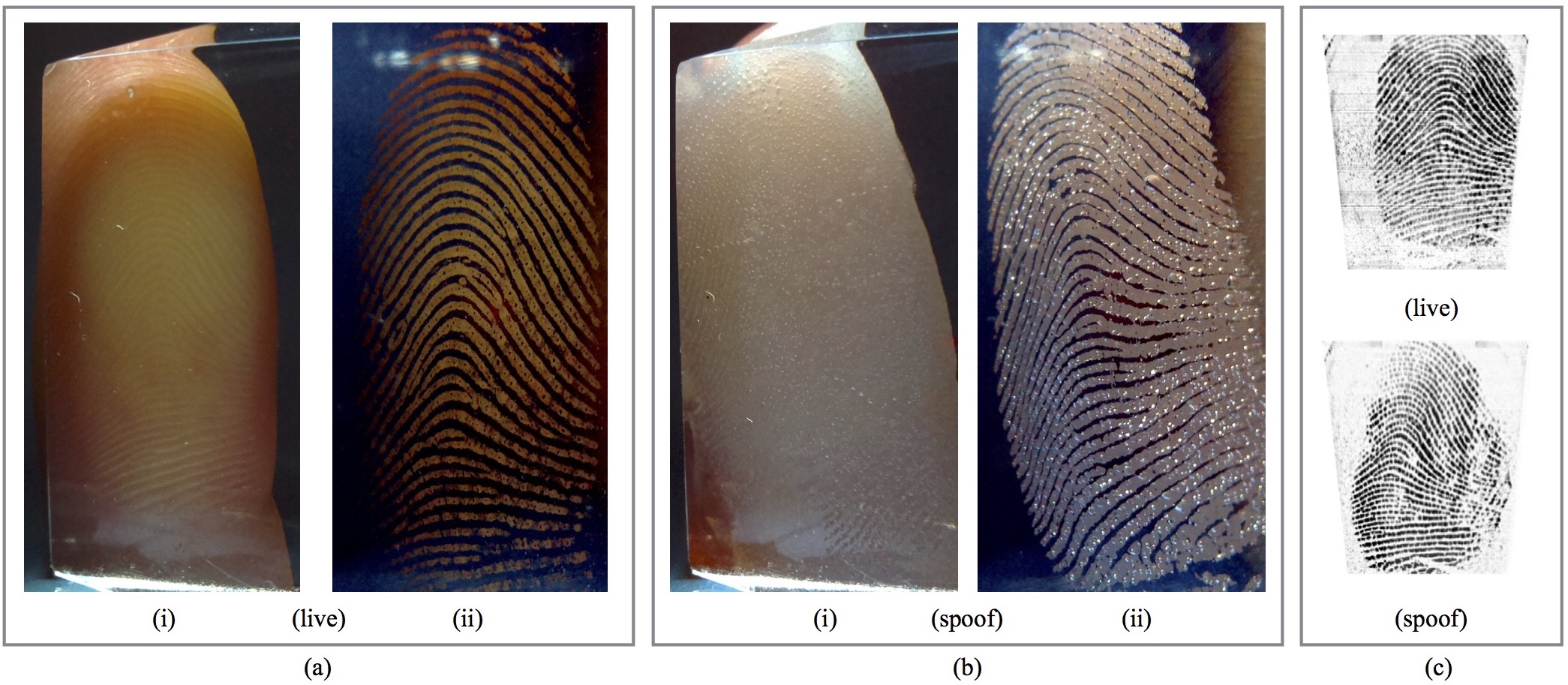}
\caption{Fingerprint images acquired using the RaspiReader. Images in (a) were collected from a live finger during a single acquisition. Images in (b) were collected from a spoof finger during a single acquisition. Using features extracted from both raw image outputs ((i), direct) and ((ii), high-contrast FTIR) of the RaspiReader, our spoof detection algorithms are able to discriminate between live fingers and spoof fingers. In particular, because the images in (i) and (ii) contain color information, discriminative color local binary patterns (CLBP) can be extracted for spoof detection. The raw FTIR image output of the RaspiReader (ii) can be post processed (after spoof detection) to output images suitable for fingerprint matching. Images in (c) were acquired from the same live finger (a) and spoof finger (b) on a COTS 500 ppi optical reader. The close similarity between the two images in (c) qualitatively illustrates why current spoof detectors are limited by the low information content, processed fingerprint images output by COTS readers.}
\vspace{-1.5em}
\end{center}
\end{figure*}

In an effort to mitigate the costs associated with presentation attacks, a number of presentation attack detection techniques involving both hardware and software have been proposed in the literature. Special hardware embedded in fingerprint readers\footnote{Several fingerprint vendors have developed hardware spoof detection solutions by employing multispectral imaging, infrared imaging (useful for sub-dermal finger analysis), and pulse capture to distinguish live fingers from spoof fingers \cite{com1, com3}.} enables capture of features such as heartbeat, thermal output, blood flow \cite{blood_flow}, odor \cite{odor}, and sub-dermal finger characteristics \cite{com1, rowe1, oct} useful for distinguishing a live finger from a spoof \cite{spoofs_survey, hardware_issues, burned, schuckers}. Presentation attack detection methods in software are based on extracting textural \cite{texture0, texture1, texture2, texture3, texture4}, anatomical \cite{pores}, and physiological \cite{perspiration1, perspiration2} features from processed\footnote{Raw fingerprint images are ``processed" (such as RGB to grayscale conversion, contrast enhancement, and scaling) by COTS readers to boost matching performance. However, useful spoof detection information (such as color and/or minute textural abberations) is lost during this processing.} fingerprint images which are used in conjunction with a classifier such as Support Vector Machines (SVM). Alternatively, a Convolutional Neural Network (CNN) can be trained to distinguish a live finger from a spoof \cite{CNN1, CNN2, tarang}. 

While hardware and software spoof detection schemes provide a reasonable starting point for solving the spoof detection problem, current solutions have a plethora of shortcomings. As noted in \cite{hardware_issues, burned, schuckers} most hardware based approaches can be easily bypassed by developing very thin spoofs (Fig. 3 (a)), since heartbeat, thermal output, and blood flow can still be read from the live human skin behind the thin spoof. Additionally, some of the characteristics (such as odor and heartbeat) acquired by the hardware vary tremendously amongst different human subjects, making it very difficult to build an adequate model representative of all live subjects \cite{hardware_issues, burned}.

\begin{figure*}[t]
\begin{center}
\includegraphics[scale=0.25]{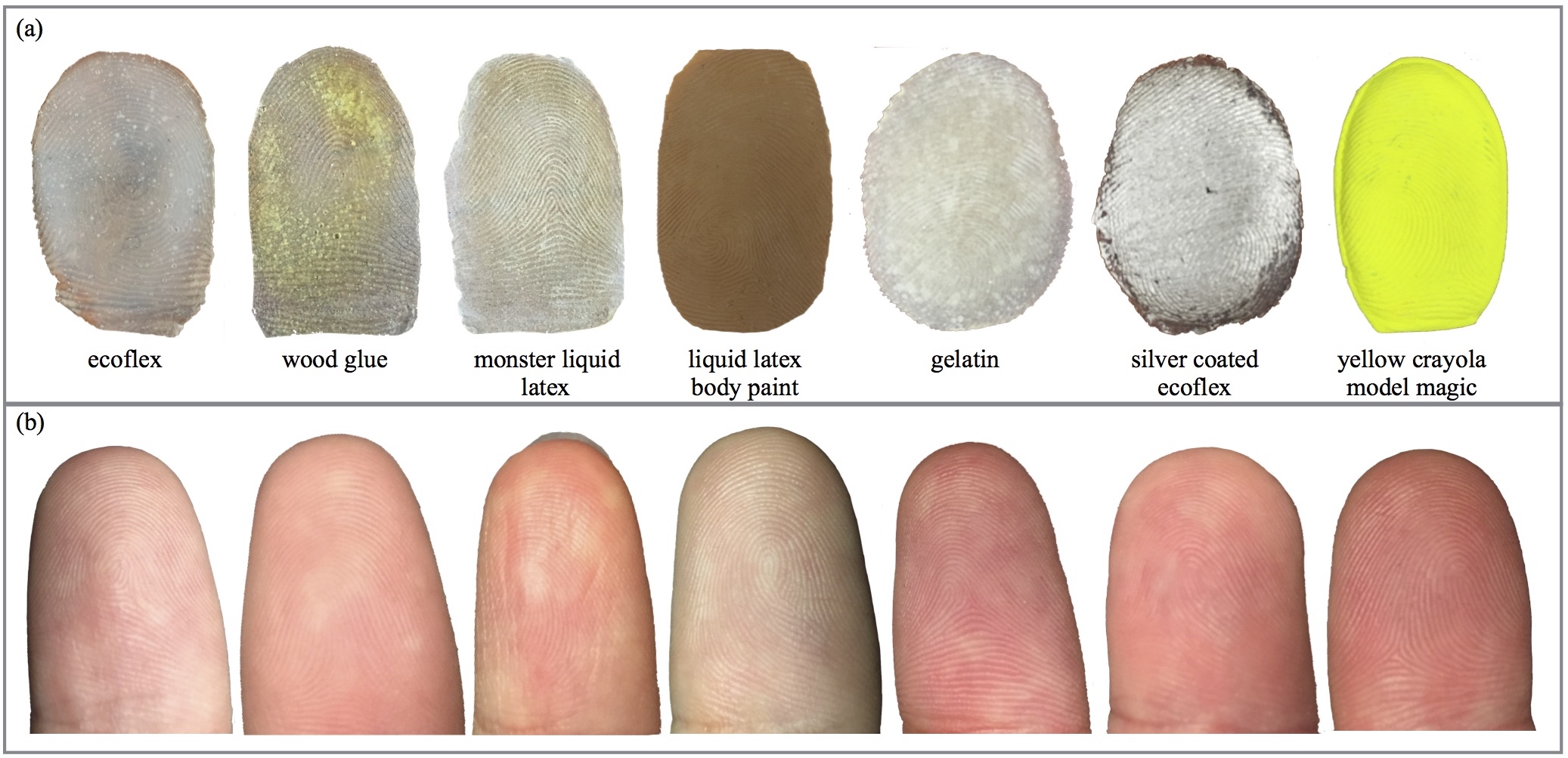}
\caption{Example spoof fingers and live fingers in our database. (a) Spoof fingers and (b) live fingers used to acquire both spoof fingerprint impressions and live fingerprint impressions for conducting the experiments reported in this paper.}
\vspace{-1.5em}
\end{center}
\end{figure*} 

Spoof detection software solutions have their own limitations. Although the LiveDet 2015 competition reported state-of-the-art spoof detection software to have an average accuracy of 95.51\% \cite{2015}, the evaluations were limited to spoofs fabricated with only six material types. Furthermore, the spoof detection performance at desired operating points such as False Detect Rate (FDR) of 0.1\% was not reported, and very limited evaluation was performed to determine the effects of testing spoof detectors with spoofs fabricated from materials not seen during training (cross-material evaluation). In the limited cross material evaluation that was performed, the rate of spoofs correctly classified as spoofs was shown to drop from 96.57\% to 94.20\% \cite{2015}. While this slight drop in accuracy seems promising, without knowing the performance at field conditions, namely False Detect Rate (FDR) of 0.1\% on a larger collection of unknown materials, the reported levels of total accuracy should be accepted with caution. Chugh et al. \cite{tarang} pushed state-of-the-art fingerprint spoof detection performance on the LiveDet 2015 dataset from 95.51\% average accuracy to 98.61\% average accuracy using a CNN, but they also demonstrated that performance at strict operating points dropped significantly in some experiments\footnote{Chugh et al. reported an average accuracy on the LiveDet 2011 dataset of 97.41\%, however, at a FDR of 1.0\%, the TDR was only 90.32\%. This indicates that current state-of-the-art spoof detection systems leave room for improvement at desired operating points.}. Additionally, many other studies have reported up to a three-fold increase in error when testing spoof detectors on unknown material types \cite{inter1,inter2,2011}. 

Because of the less than desired performance of spoof detection software to adapt to spoofs fabricated from unseen materials, studies in \cite{open1}, \cite{open2}, and \cite{open3} developed open-set recognition classifiers to better detect spoofs fabricated with novel material types. However, while these classifiers are able to generalize to spoofs made with new materials better than closed-set recognition algorithms, their overall accuracy (approx. 85\% - 90\%) still does not meet the desired performance for field deployments. 

Given the limitations of state-of-the-art fingerprint presentation attack detection (both in hardware and software), it is evident that much work remains to be done in developing robust and generalizable presentation attack detection solutions. We posit that one of the biggest limitations facing the most successful spoof detection solutions to date (such as use of textural features \cite{2011} and CNNs \cite{CNN1, CNN2, tarang}), is the processed COTS fingerprint reader images used to train spoof detectors. In particular, because COTS fingerprint readers output fingerprint images which have undergone a number of image processing operations (in an effort to achieve high matching performance), they are not optimal for fingerprint spoof detection, since valuable information such as color and textural aberrations is lost during the image processing operations. By removing color and minute textural details from the raw fingerprint images, spoof fingerprint impressions and live fingerprint impressions (acquired on COTS optical readers) become very similar (Fig. 2 (c)), even when the physical live/spoof fingers used to collect the respective fingerprint impressions appear very different (Fig. 3). 

This limitation inherent to many existing spoof detection solutions motivated us to develop a custom, optical fingerprint reader, called RaspiReader, with the capability to output 2 raw images (from 2 different cameras) for spoof detection. By mounting two cameras at appropriate angles to a glass prism (Fig. 4), one camera is able to capture high contrast FTIR fingerprint images (useful for both fingerprint spoof detection and fingerprint matching) (Fig. 2 (ii)), while the other camera captures direct images of the finger skin in contact with the platen (useful for fingerprint spoof detection) (Fig. 2 (i)). Both images of the RaspiReader visually differentiate between live fingers and spoof fingers much more than the processed fingerprint images output by COTS fingerprint readers (Fig. 2 (c)). 

\newcommand{\specialcell}[2][c]{%
  \begin{tabular}[#1]{@{}c@{}}#2\end{tabular}}
  \newcommand{\tabitem}{~~\llap{\textbullet}~~}

 \begin{table*}[t]
 \centering
\begin{threeparttable}
\resizebox{\textwidth}{!}{%
\begin{tabular}{ |c||M{5.5cm}|c|c|}
 \multicolumn{4}{c}{{\tiny Table 1: Primary Components Used to Construct RaspiReader. Total Cost is \$165.48}} \\
 \hline
{\tiny Component Image} & {\tiny Name and Description} & {\tiny Quantity} & {\tiny Cost (USD)\tnote{1}}\\
\hline
 \includegraphics[valign=m,scale=0.10]{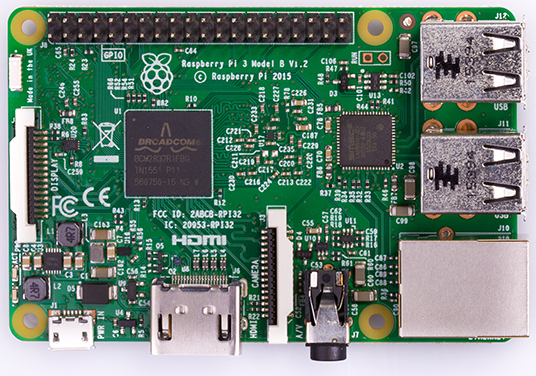} & {\tiny \textbf{Raspberry Pi:}  A single board computer (SBC) with 1.2 GHz 64-bit quad-core CPU, 1 GB RAM, MicroSDHC storage, and Broadcom VideoCore IV Graphic card} & {\tiny1} & {\tiny \$38.27}\\
 \hline
 \includegraphics[valign=m,scale=0.075]{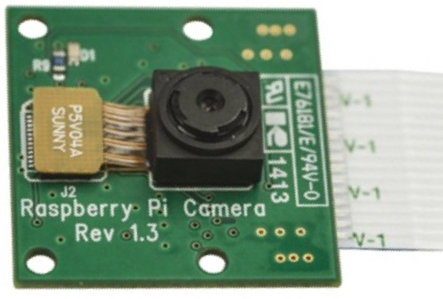}  & {\tiny \textbf{Raspberry Pi Camera Module V1:} A 5.0 megapixel, 30 frames per second, fixed focal length camera} & {\tiny 2} & {\tiny \$22.44}\\
  \hline
 \includegraphics[valign=m,scale=0.20]{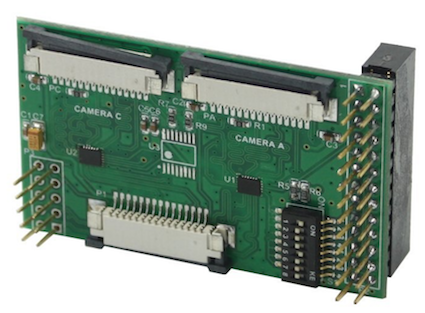} & {\tiny \textbf{Multi-Camera Adapter:} Splits Raspberry Pi camera slot into two slots, enabling connection of two cameras} & {\tiny 1} & {\tiny \$49.99}\\
  \hline
 \includegraphics[valign=m,scale=0.10]{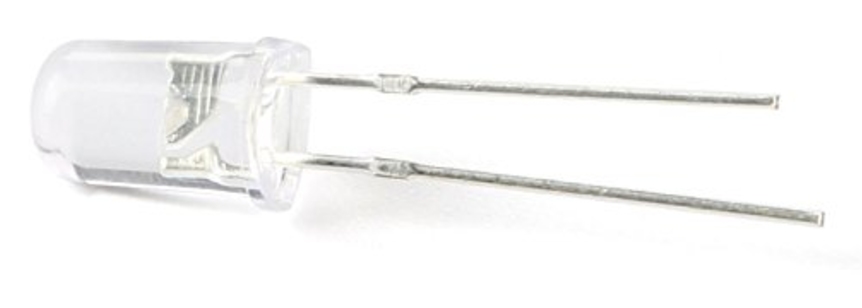}  & {\tiny \textbf{LEDs:} white light, 5 mm, 1 watt} & {\tiny3} & {\tiny \$0.10}\\
  \hline
\includegraphics[valign=m,scale=0.15]{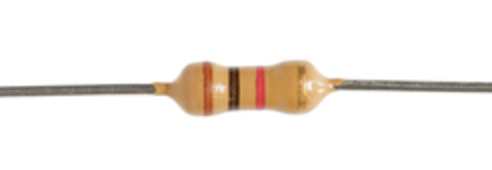} & {\tiny \textbf{Resistors:} 10 k$\Omega$} & {\tiny3} & {\tiny \$5.16}\\
  \hline
\includegraphics[valign=m,scale=0.15]{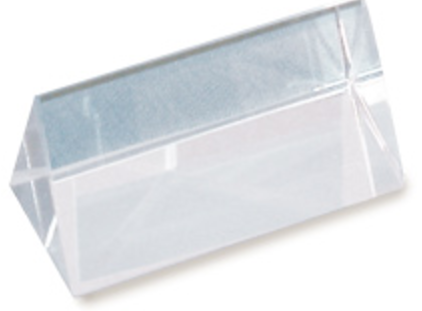}  & {\tiny \textbf{Right Angle Prism:} Acrylic, 25 mm} & {\tiny 1} & {\tiny \$16.56}\\
\hline
 
\end{tabular}}
\begin{tablenotes}
\item[1] All items were purchased for the listed prices on Amazon.com
\end{tablenotes}
\end{threeparttable}
\vspace{-1.0em}
\end{table*} 

RaspiReader's two camera approach is similar to that which was prescribed by Rowe et al. in \cite{com1, rowe1} where both an FTIR image and a direct view image were acquired using different wavelength LEDs, however, the commercial products developed around the ideas in \cite{com1, rowe1} act as a proprietary black box outputting only a single processed composite image of a collection of raw image frames captured under various wavelengths. As such, fingerprint researchers cannot implement new spoof detection schemes on the individual raw frames captured by the reader. Furthermore, unlike the patented ideas in \cite{com1}, RaspiReader is built with ubiquitous components and open source software packages, enabling fingerprint researchers to very easily prototype their own RaspiReader, and further customize it with new spoof detection hardware.

By utilizing the novel hardware of the RaspiReader, we are able to develop robust fingerprint presentation attack detection algorithms. In particular, because both image outputs of the RaspiReader are raw and contain useful color information, we can extract discriminative color local binary patterns (CLBP) \cite{color_lbp_rec, color_lbp_spoofing} from each of the image outputs. Our experimental results demonstrate that the color local binary patterns from each image contain complementary information such that when the features are fused together and passed to a binary SVM classifier, state-of-the-art spoof detection performance can be achieved. Additionally, by calibrating and processing the FTIR image output of the RaspiReader (post spoof detection), we demonstrate that RaspiReader is not only interoperable with existing COTS optical readers but is also capable of achieving state-of-the-art fingerprint matching accuracy. Finally, RaspiReader is cost effective (approx. \$165 in total component costs), built with off-the-shelf components (Table 1), and can be easily assembled and operated by those with little expertise in hardware, enabling easy integration and extension of this research (such as adding additional hardware specifically for spoof detection).

More concisely, the contributions of this research are:

\begin{itemize}
\item An open source, easy to assemble, cost effective fingerprint reader, called RaspiReader, capable of producing fingerprint images useful for spoof detection and that have sufficient quality for fingerprint matching. The custom RaspiReader can be easily modified to facilitate presentation attack detection studies.

\item A customized fingerprint reader with two cameras for image acquisition rather than a single camera. Use of two cameras enables robust fingerprint spoof detection, since we can extract features from two complementary, information rich images instead of processed grayscale images output by traditional COTS optical fingerprint readers.
  
  \item The implementation of a new fingerprint spoof detection schema through the extraction of highly discriminative fingerprint spoof detection features (CLBP) using the raw images of the RaspiReader. Our algorithm delivers state-of-the-art spoof detection accuracy.
  
\end{itemize}

\begin{figure}[h]
\begin{center}
\includegraphics[scale=0.21]{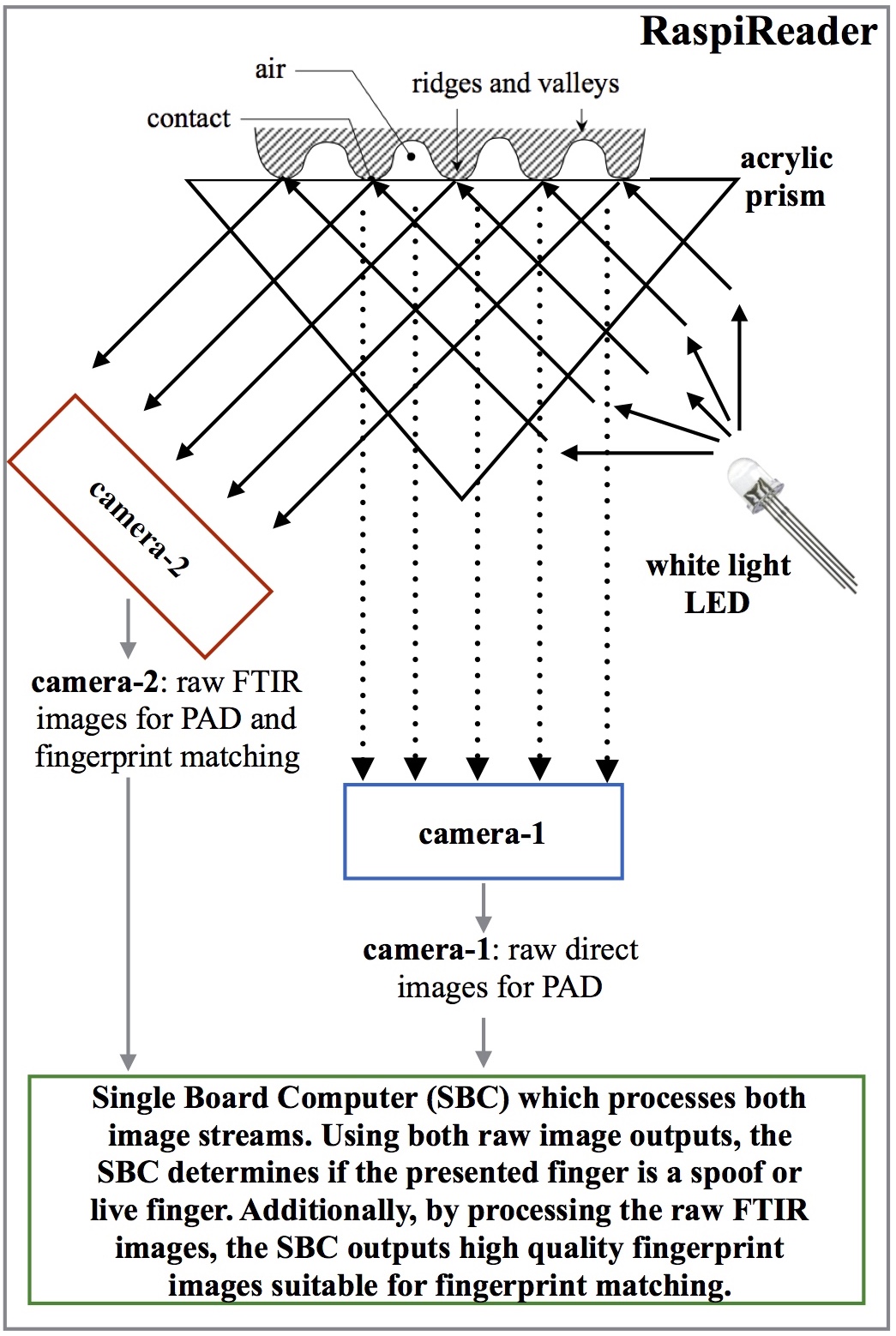}
\caption{Schematic illustrating RaspiReader functionality. Incoming white light from three LEDs enters the prism. Camera 2 receives light rays reflected from the fingerprint ridges only (light rays are not reflected back from the valleys due to FTIR). This image from Camera 2, with high contrast between ridges and valleys can be used for both spoof detection and fingerprint matching. Camera 1 receives light rays reflected from both the ridges and valleys. This image from Camera 1 provides complementary information for spoof detection.}
\vspace{-1.5em}
\end{center}
\end{figure} 

\section{RaspiReader Construction and Calibration}

In this section, the construction of the RaspiReader using ubiquitous, off-the-shelf components is explained. Additionally, the steps for calibrating and processing the raw FTIR fingerprint images of the RaspiReader for fingerprint matching are prescribed.

\subsection{Construction of RaspiReader}

The construction of RaspiReader adheres to the following process. First, an outer casing is electronically modeled using Meshlab (Fig. 5) \cite{meshlab} and subsequently fabricated using a high resolution 3D printer \cite{printer}. Next, the components enumerated in Table 1 are assembled together according the the specifications of the schematic diagram shown in Figure 4. The exact dimensions of assembly are not critical, however, depending on the dimensions used, the focal lengths of the two cameras need to be adjusted to focus on the finger surface as it makes contact with the acrylic prism.

Because the Raspberry Pi only has a single camera connection port, a camera port multiplexer is used to enable the use of multiple cameras on a single Pi \cite{multi}. Using the Raspberry Pi GPIO pins, the code available in \cite{multi}, and the camera multiplexer, one can easily extend the Raspberry Pi to use multiple cameras. Note that an alternative approach that has been suggested for enabling multiple cameras on the Raspberry Pi is to attach one camera to the Pi's camera connection port, and a separate USB web-cam to the Raspberry Pi USB port. This method was experimented with, however, the frame rate of the USB camera is significantly reduced on the Pi due to the latency in loading images from the USB port to the Pi's graphics card. As such, using the camera multiplexer is recommended. 

Once the components of Table 1 have been assembled, and the camera port multiplexed for two cameras, open source python libraries \cite{multi} can be used to acquire two images from the fingerprint reader (one raw FTIR fingerprint image and another raw direct fingerprint image). 

\begin{figure}[t]
\begin{center}
\includegraphics[scale=0.25]{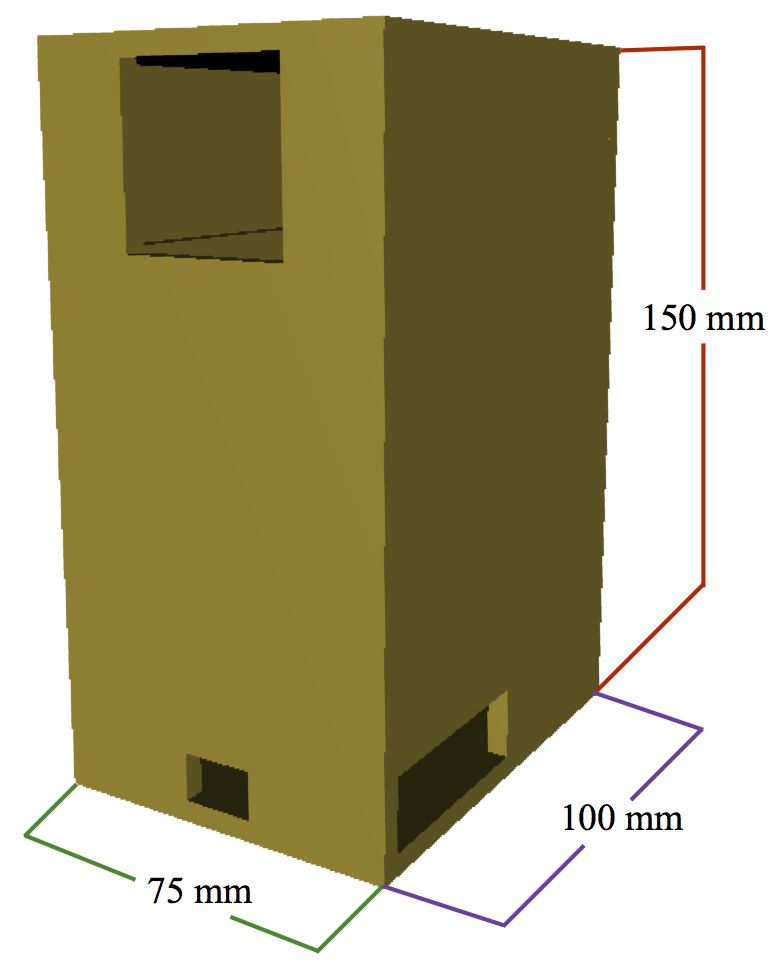}
\caption{Electronic CAD model of the RaspiReader case. The dimensions here were provided to a 3D printer for fabricating the prototype.}
\vspace{-1.0em}
\end{center}
\end{figure} 

\begin{figure}[!h]
\begin{center}
\includegraphics[scale=0.25]{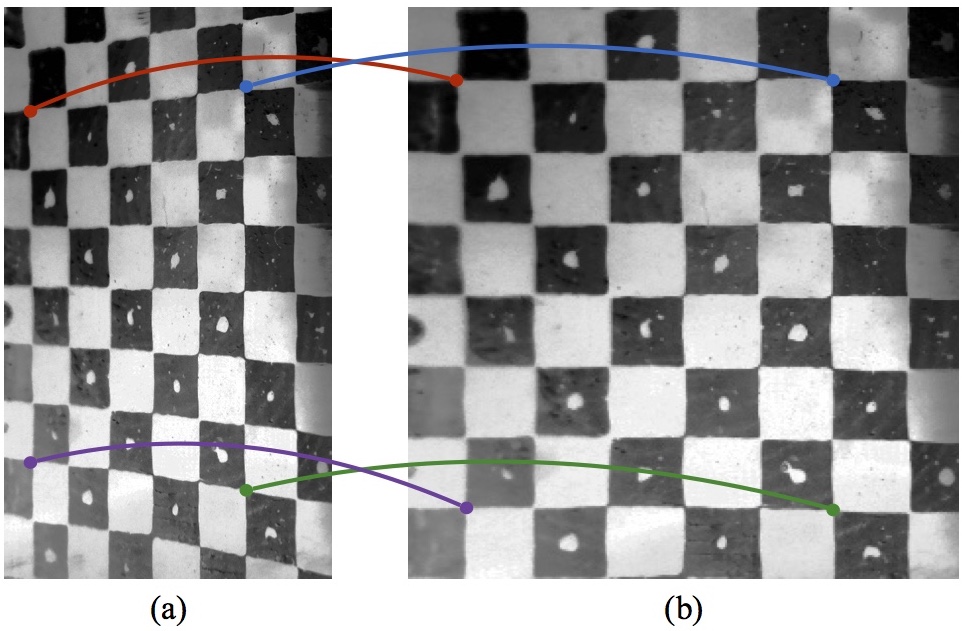}
\caption{Acquiring Image Transformation Parameters. (a) A 3D printed checkerboard pattern is imaged by the RaspiReader. In this image, the checkerboard pattern is not frontalized to the image plane. Therefore, destination coordinate pairs are defined on a frontalized checkerboard (b) such that perspective transformation parameters can be estimated to map (a) into (b). These transformation parameters are subsequently used to frontalize fingerprint images acquired by RaspiReader for the purpose of fingerprint matching. }
\vspace{-1.5em}
\end{center}
\label{intro_fig}
\end{figure} 

\begin{figure*}[t]
\begin{center}
\includegraphics[scale=.3]{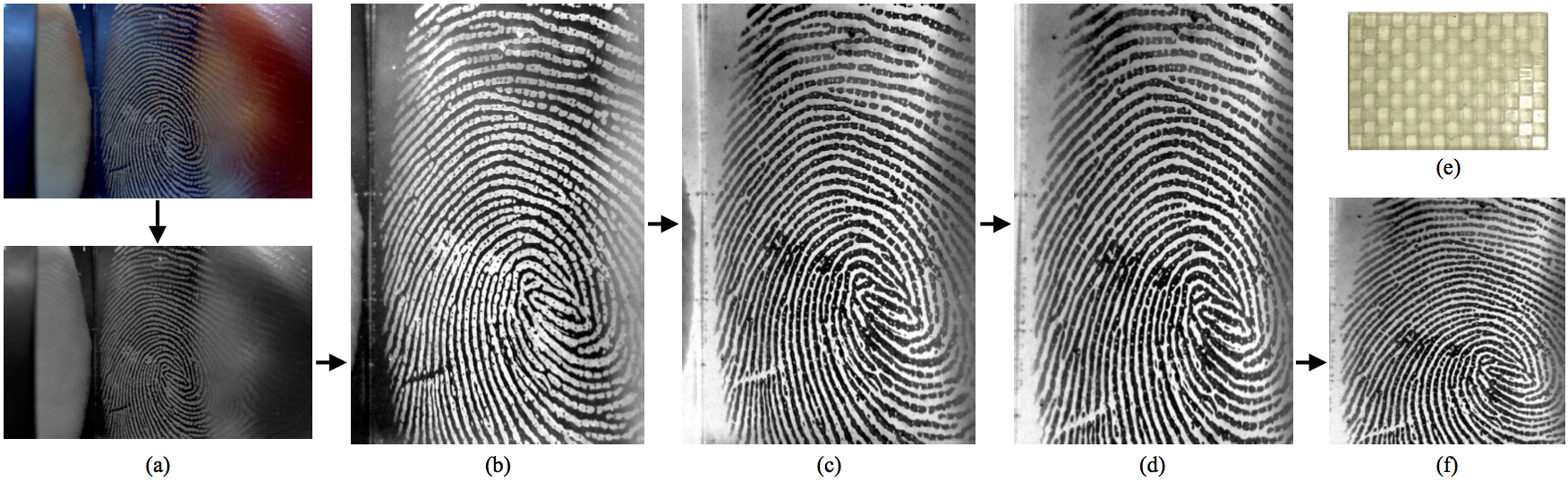}
\caption{Processing a RaspiReader raw FTIR fingerprint image into a 500 ppi fingerprint image compatible for matching with existing COTS fingerprint readers. (a) The RGB raw FTIR image is first converted to grayscale. (b) Histogram equalization is performed on the grayscale FTIR image to enhance the contrast between the fingerprint ridges and valleys. (c) The fingerprint is negated so that the ridges appear dark, and they valleys appear white. (d),(f) Calibration (estimated using the checkerboard calibration pattern in (e)) is applied to frontalize the fingerprint image to the image plane and down sample (by averaging neighborhood pixels) to 500 ppi in both the x and y directions.}
\vspace{-1.5em}
\end{center}
\end{figure*} 

\subsection{Fingerprint Image Processing}

In order for the RaspiReader to be used for spoof detection, it must also demonstrate the ability to output high quality fingerprint images suitable for fingerprint matching. As previously mentioned, the RaspiReader performs spoof detection on non-processed, raw fingerprint images. While these raw images are shown to provide discriminatory information for spoof detection, they need to be made compatible with processed images output by other commercial fingerprint readers. Therefore, after spoof detection, the RaspiReader performs (1) image enhancement operations and (2) image transformations on the raw high contrast, FTIR image frames in order to output high fidelity images compatible with COTS optical fingerprint readers (Fig. 7).  

\subsubsection{Fingerprint Image Enhancement}

Let a raw (unprocessed) FTIR fingerprint image from the RaspiReader be denoted as $FTIR_{raw}$. This raw image $FTIR_{raw}$ is first converted from the RGB color space to grayscale ($FTIR_{gray}$) (Fig. 7 (a)). Then, in order to further contrast the ridges from the valleys of the fingerprint, histogram equalization is performed on $FTIR_{gray}$ (Fig. 7 (b)). Finally, $FTIR_{gray}$ is negated so that the ridges of the fingerprint image are dark, and the background of the image is white (as are fingerprint images acquired from COTS readers) (Fig. 7 (c)). 

\subsubsection{Fingerprint Transformation}

Following the aforementioned image processing techniques, the RaspiReader FTIR fingerprint images are further processed by performing a perspective transformation (to frontalize the fingerprint to the image plane) and scaling to 500 ppi (the native resolution of the RaspiReader images, approx. 900 ppi, was obtained using a calibration target.) (Fig. 7 (d),(f)).

A perspective transformation is performed using Equation 1,

\begin{equation}
\begin{bmatrix}x'\\ y'\\ 1\end{bmatrix}=\frac{1}{\lambda }\begin{bmatrix} a & b & c\\ d& e & f\\ g & h & 1\end{bmatrix}\begin{bmatrix}x\\ y\\ 1\end{bmatrix}
\end{equation} where $x$ and $y$ are the source coordinates, $x'$ and $y'$ are the transformed coordinates, $(a, b, c, d, e, f, g, h)$ is the set of transformation parameters, and $\lambda = gx+hy+1$ is a scale parameter. In this work, we image a 3D printed checkerboard pattern (Fig. 7 (e)) to define source and destination coordinate pairs (Fig. 6) such that the transformation parameters could be estimated. Once the perspective transformation has been completed, the image is downsampled (by averaging neighborhood pixels) to 500 ppi (Fig. 7 (f)).

Upon completion of this entire fingerprint reader assembly and image processing procedure, the RaspiReader is fully functional and ready for use in both presentation attack detection and subsequent fingerprint matching.

\section{Live and Spoof Fingerprint Database Construction}

To test the utility of the RaspiReader for presentation attack detection and its interoperability for fingerprint matching, a database of live and spoof fingerprint impressions was collected for performing experiments. This database is constructed as follows. 

Using 7 different materials (Fig. 3 (a)), 66 spoofs were fabricated. Then, for each of these spoofs, 10 impressions were captured at varying orientations and pressure on both the RaspiReader (Rpi) and a COTS 500 ppi optical FTIR fingerprint reader ($COTS_A$). The summary of this data collection is enumerated in Table 2.

To collect a sufficient variety of live finger data, we enlisted 15 human subjects with different skin colors (Fig. 3 (b)). Each of these subjects gave 5 finger impressions (at different orientations and pressures) from all 10 of their fingers on both the RaspiReader and $COTS_A$ \footnote{Acquiring a fingerprint on RaspiReader involves the same user interactions that a COTS optical reader does. A user simply places their finger on an acrylic prism. Then, LEDs illuminate the finger surface and images are captured from both cameras over a time period of 1 second (Fig. 1). The only difference in the acquisition process between a COTS reader and RaspiReader is that RaspiReader acquires two complementary images of the finger in contact with the acrylic platen from two separately mounted cameras.}. A summary of this data collection is enumerated in Table 3.

\begin{table}[h]
 \centering
\begin{threeparttable}
\begin{tabular}{ |c||c|c|c|c|}
 \multicolumn{5}{c}{Table 2: Summary of Spoof\tnote{1} Fingerprints Collected} \\
 \hline
 Material & \specialcell{Number \\of \\Spoofs\tnote{2}} &\specialcell{RPi Direct \\Images} & \specialcell{RPi FTIR \\Images} & \specialcell{$COTS_A$ \\FTIR \\Images} \\
 \hline
 \hline
 Ecoflex & 10 & 100 & 100 & 100\\
 \hline
 \specialcell{Wood Glue} & 10 & 100 & 100 & 100\\
  \hline
 \specialcell{Monster Liquid \\Latex} & 10 & 100 & 100 & 100\\
 \hline
 \specialcell{Liquid Latex \\Body Paint} & 10 & 100 & 100 & 100\\
 \hline
 Gelatin & 10 & 100 & 100 & 100\\
 \hline
 \specialcell{Silver Coated\\ Ecoflex} & 10 &  100 & 100 & 100\\
 \hline
 \specialcell{Crayola Model\\ Magic} & 6 & 60 & 60 & 60\\
 \hline
  \hline
  Total & 66 & 660 & 660 & 660\\
  \hline
\end{tabular}
\begin{tablenotes}
\item[1] The spoof materials used to fabricate these spoofs were in accordance with the approved materials by the IARPA Odin project \cite{odin}.
\item[2] The spoofs are all of unique fingerprint patterns.
\end{tablenotes}
\end{threeparttable}
\end{table}

\begin{table}[h]
 \centering
\begin{tabular}{ |c|c|c|c|c|}
 \multicolumn{5}{c}{Table 3: Summary of Live Finger Data Collected} \\
 \hline
 \specialcell{Number of \\Subjects} & \specialcell{Number of \\Fingers} &  \specialcell{RPi Direct \\Images} &\specialcell{RPi FTIR \\Images} & \specialcell{$COTS_A$ \\FTIR\\Images} \\
 \hline
 \hline
 15 & 150 & 750 & 750 & 750\\
 \hline
\end{tabular}
\end{table}

In addition to the images of live finger impressions and spoof finger impressions we collected for conducting spoof detection experiments, we also verified that for spoofs with optical properties and/or 3D structural properties too far from that of live finger skin (Fig. 8), images would not be captured by the RaspiReader. These ``failure to capture" spoofs are therefore filtered out as attacks before any software based spoof detection methods need to be performed.

\begin{figure}[!h]
\begin{center}
\includegraphics[scale=0.225]{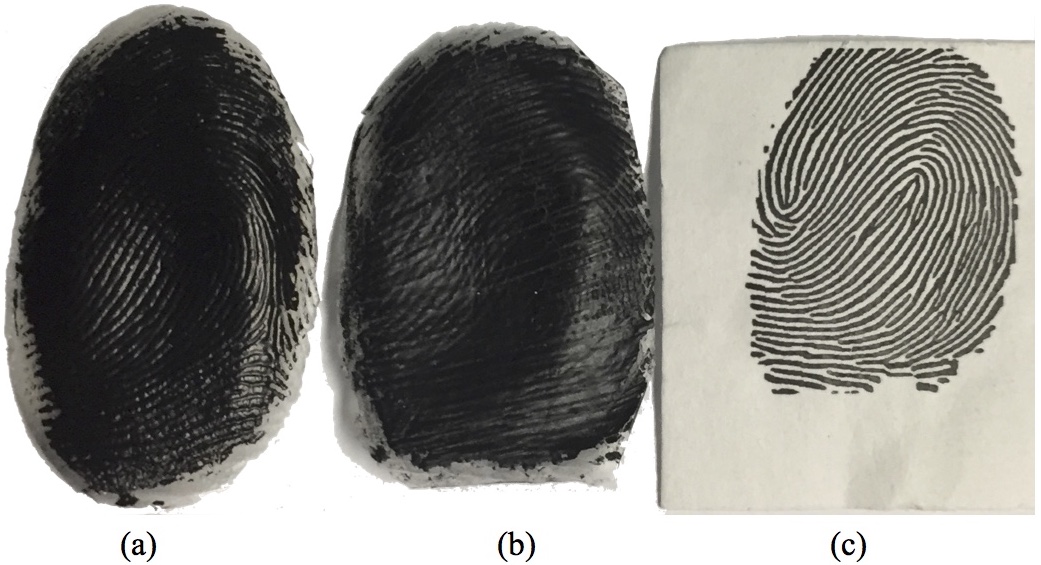}
\caption{Failure to Capture. Several spoofs are unable to be imaged by the RaspiReader due to their dissimilarity in color (a, b) and/or lack of 3D friction ridge patterns (c). In particular, because (a) and (b) are black, all light rays will be absorbed preventing light rays from reflecting back to the FTIR imaging sensor. In (c), the lack of depth on the friction ridge patterns will again prevent the FTIR phenomena. (a) and (b) are both ecoflex spoofs coated with two different conductive coatings. (c) is a 2D printed spoof attack. }
\vspace{-1.5em}
\end{center}
\label{intro_fig}
\end{figure} 

\section{Experimental Protocol and Results}

Given the database of live and spoof fingerprint images collected on both $COTS_A$, and the prototype RaspiReader, a number of spoof detection experiments are conducted to demonstrate the value of the raw images from the RaspiReader for training spoof detectors. In addition, experiments are conducted to demonstrate that fingerprint images from the RaspiReader are compatible for matching with fingerprint images acquired from $COTS_A$.

\subsection{Spoof Detection Experiments}

In all of our spoof detection experiments, we use a binary linear SVM for classification\footnote{Other classifiers were experimented with such as K-Nearest Neighbor and Support Vector Machines with various kernels (RBF and Polynomial), however, linear SVM provided the best results in our experiments.}, namely to classify images into live vs. spoof fingerprints. A binary linear SVM informally is the hyperplane with the ``largest" margin separating the positive and negative samples (here spoof and live impressions). More formally, this optimal decision boundary can be found by solving the following unconstrained optimization equation (Eq. 2) given a set of instance label pairs $(\mathbf{x}_i, y_i), i = 1,...,n, \mathbf{x}_i \in R^{d}, y_i \in \{-1, +1\}$ and the quadratic hinge loss function $\xi^2 (\mathbf{w};\mathbf{x}_i;y_i)$ defined in Equation 3 \cite{two_class}.

\begin{equation}
\min_\mathbf{w} \frac{1}{2}\mathbf{w}^{T}\mathbf{w}+C\sum_{i=1}^{n}\xi^2 (\mathbf{w};\mathbf{x}_i;y_i) 
\end{equation}

\begin{equation}
\xi^2 (\mathbf{w};\mathbf{x}_i;y_i)=\max(0,1-y_i\cdot \mathbf{w}\cdot (\phi (\mathbf{x}_i)+ b))^2 
\end{equation}

In practice, finding a ``large margin" (hyperplane) which perfectly separates the positive and negative samples is infeasible due to noise in the training data. As such, linear SVMs are relaxed from a ``hard margin" SVM (no training sample misclassifications allowed by the margin) to a ``soft margin" SVM (some training sample misclassifications allowed by the margin). The number of misclassifications allowed is dependent upon the user defined parameter $C$ in Equation 2. If $C$ is set very large, then a large penalty incurs for allowing any training sample misclassifications. Consequently, a very strict decision boundary is chosen which may not generalize well to the testing data. In an effort to better generalize the SVM to the testing data, the value of parameter $C$ can be reduced. However, if $C$ is reduced too much, the selected decision boundary may begin to deviate from the optimal decision boundary for the testing data. Thus, selecting the appropriate value of $C$ is a tradeoff that needs to be empirically determined for the dataset on hand. In our case, we use five-fold cross validation to select the value of $C$ (from the list of $\begin{bmatrix} 10^{-5} & 10^{-4} & ... & 10^4 & 10^5\end{bmatrix}$) such that the best performance is achieved in different folds. In our experiments, the best classification results were achieved with $C = 10^2$.

{\it Experiment 1: LBP Features From COTS Acquired Images.} With a classifier selected, we begin our experiments using grayscale processed fingerprint images acquired from $COTS_A$ (Fig. 9) for training and testing. The purpose of this initial experiment is to set a baseline performance for spoof detection when classifiers are trained on grayscale images from COTS fingerprint readers. In the subsequent experiments, the images from the RaspiReader are used for training classifiers. This first experiment is then used as a reference point in demonstrating the improvements in classification error when RaspiReader images are used for extracting features rather than images from a 500 ppi COTS optical fingerprint reader. 

\begin{figure}[t]
  \centering
  \subfloat[]{\includegraphics[scale=.345]{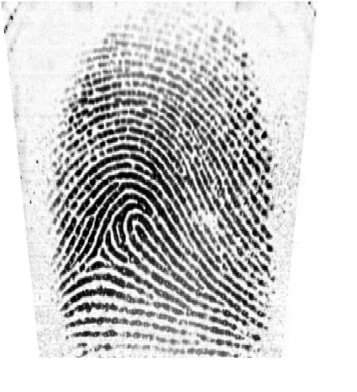}\label{fig:f1}}
  \hfill
  \subfloat[]{\includegraphics[scale=2.4]{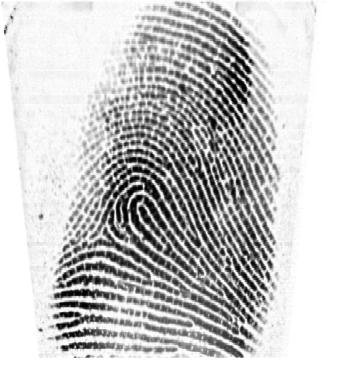}\label{fig:f2}}
  \caption{Fingerprint images acquired using a 500 ppi COTS optical fingerprint reader used to conduct a baseline spoof detection experiment. A live fingerprint impression is shown in (a). An ecoflex spoof fingerprint impression is shown in (b).}
\end{figure}

Given the reported success of local textural features for spoof detection in comparison to anatomical or physiological features \cite{2011}, we use the very prevalent grayscale and rotation invariant local binary patterns (LBP) as features for this baseline experiment \cite{g_lbp}. LBP features are extracted by constructing a histogram of bit string values determined by thresholding pixels in the local neighborhoods around each pixel in the image. Since textures can be present at different spatial resolutions, parameters $R$ and $P$ are specified in LBP construction to indicate the length (in pixels) of the neighborhood radius used for selecting pixels and also the number of neighbors to consider in a local neighborhood. Previous studies have shown that more than 90\% of fundamental textures in an image can belong to a small subset of binary patterns called ``uniform" textures (local binary patterns containing two or fewer 0/1 bit transitions) \cite{g_lbp}. Therefore, in line with previous studies using local binary patterns for fingerprint spoof detection, we also employ the use of uniform local binary patterns. More formally, extracting a uniform local binary pattern for a pixel $g_c$ and a set of $P$ neighborhood pixels $\begin{bmatrix}g_0 & g_1 & ...  & g_{P-1}& g_P\end{bmatrix}$ selected at radius $R$ from $g_c$ (Fig. 10) as defined in \cite{g_lbp} is performed by:

\begin{equation}
LBP_{P,R} = \begin{cases} 
\sum_{p=0}^{P-1}s(g_p-g_c), & \text{if } U(LBP_{P,R}) \leq 2 \\
P+1, & \text{otherwise}
\end{cases}
\end{equation}

where, 

\begin{multline}
U(LBP_{P,R}) = \mid s(g_{P-1} - g_c) - s(g_0 - g_c) \mid \\+ \sum_{p=1}^{P-1} \mid s(g_p - g_c) - s(g_{p-1} - g_c) \mid
\end{multline}

and

\begin{equation}
s(x) = \begin{cases} 
1 & \text{if }x\geq 0 \\
0 & \text{if }x< 0
\end{cases}
\end{equation}

\begin{figure}[h]
  \centering
\includegraphics[scale=.35]{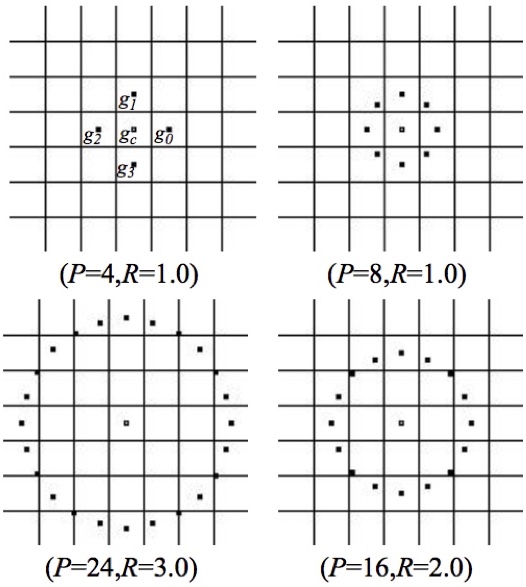}
  \caption{Circularly symmetric isotropic neighbor sets for different values of (P, R), used for the extraction of LBP. Figure reproduced from \cite{g_lbp}.}
\end{figure}

Finally, let $H_{P,R}$ be the uniform local binary pattern histogram constructed by binning the local binary patterns for each pixel $g_c$ in an image according to equations 4, 5, and 6 using $P$ and $R$ as parameters to the LBP feature extractor. In our experiments, we extract $H_{8,1}$, $H_{16,2}$, and $H_{24,3}$ in order to capture textures at different spatial resolutions. These histograms (each having $P + 2$ bins) are individually normalized and concatenated into a single feature vector $\mathbf{x}$ of dimension 54.

Using the live and spoof finger images from $COTS_A$ as described in Tables 2 and 3, five-fold cross validation is used to partition training and testing splits in the data. When partitioning each fold of the live fingers, the finger impressions from 12 subjects are used for training, and the finger impressions from the remaining 3 subjects are used for testing. When partitioning spoof finger impressions, 80\% of the impressions for each spoof material are used for training, and the other 20\% for testing.

Next, a feature vector $\mathbf{x}$ is extracted from every image in the training and testing splits and used to train and test a binary linear SVM according to the parameters previously described. Using the proposed classification scheme and training/testing splits, we report a mean (over five folds) True Detection Rate (TDR) of 58.33\% at a False Detection Rate (FDR) of 0.1\%. Additionally, we report the standard deviation for the TDR (over five folds) to be 37.69\%. 

Not only is the performance of this baseline classifier unacceptably low, it is also not robust, as evidenced by the extremely high standard deviation of the TDR over the five folds. Furthermore, it demonstrates that 500 ppi processed grayscale fingerprint images from COTS FTIR fingerprint readers have relatively low information content for discriminating spoof fingers from live fingers. This serves as motivation for developing the RaspiReader to output two raw RGB images for spoof detection. In the subsequent sections, we perform experiments on the RaspiReader images and demonstrate that the performance of classifiers trained with RaspiReader images far surpass the performance of classifiers trained on processed grayscale images.
 
{\it Experiment 2: Color LBP Features Extracted from RaspiReader Raw FTIR Images.} In this experiment, we make use of the information rich, raw FTIR images from the RaspiReader (Fig. 11) for presentation attack detection. As with the experiment 1, we again pursue the use of LBP textural features given their superior performance in comparison with other types of features that have been tried for fingerprint spoof detection \cite{2011}. However, since the raw FTIR images from the RaspiReader contain color information, rather than using the traditional grayscale LBP features, we employ the use of color local binary patterns (CLBP). Previous works have shown the efficacy of CLBP for both face recognition and face spoof detection \cite{color_lbp_rec, color_lbp_spoofing}. However, because fingerprint images from COTS fingerprint readers are grayscale, CLBP features have, to our knowledge, not been investigated for use in fingerprint spoof detection until now. 

\begin{figure}[t]
  \centering
  \subfloat[]{\includegraphics[scale=.225]{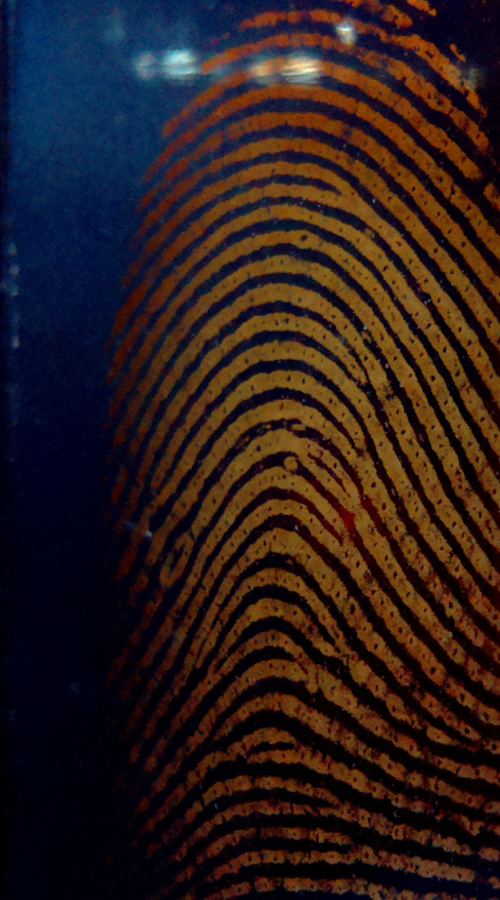}\label{fig:f1}}
  \hfill
  \subfloat[]{\includegraphics[scale=.225]{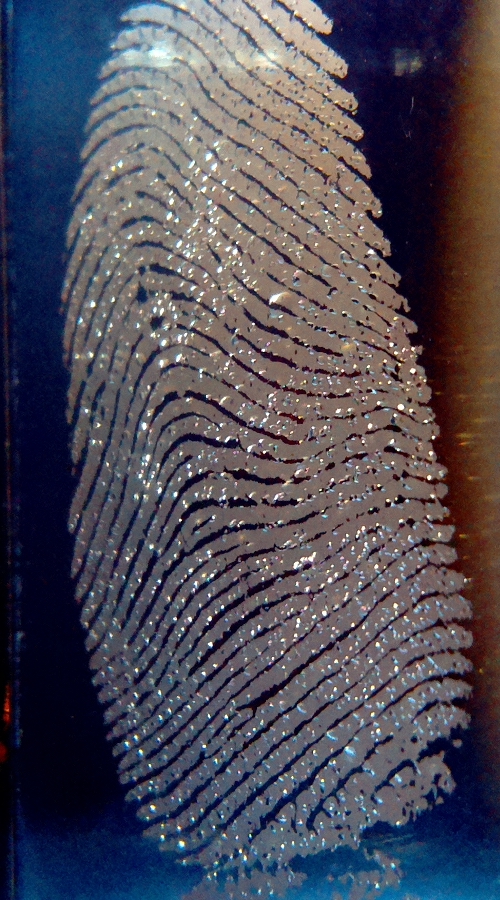}\label{fig:f2}}
  \caption{Raw FTIR fingerprint images acquired using the RaspiReader fingerprint reader and used to conduct spoof detection Experiment 2. A live fingerprint impression is shown in (a). An ecoflex spoof fingerprint impression is shown in (b).}
  \vspace{-1.5em}
\end{figure}

\begin{figure*}[t]
\begin{center}
\includegraphics[scale=0.25]{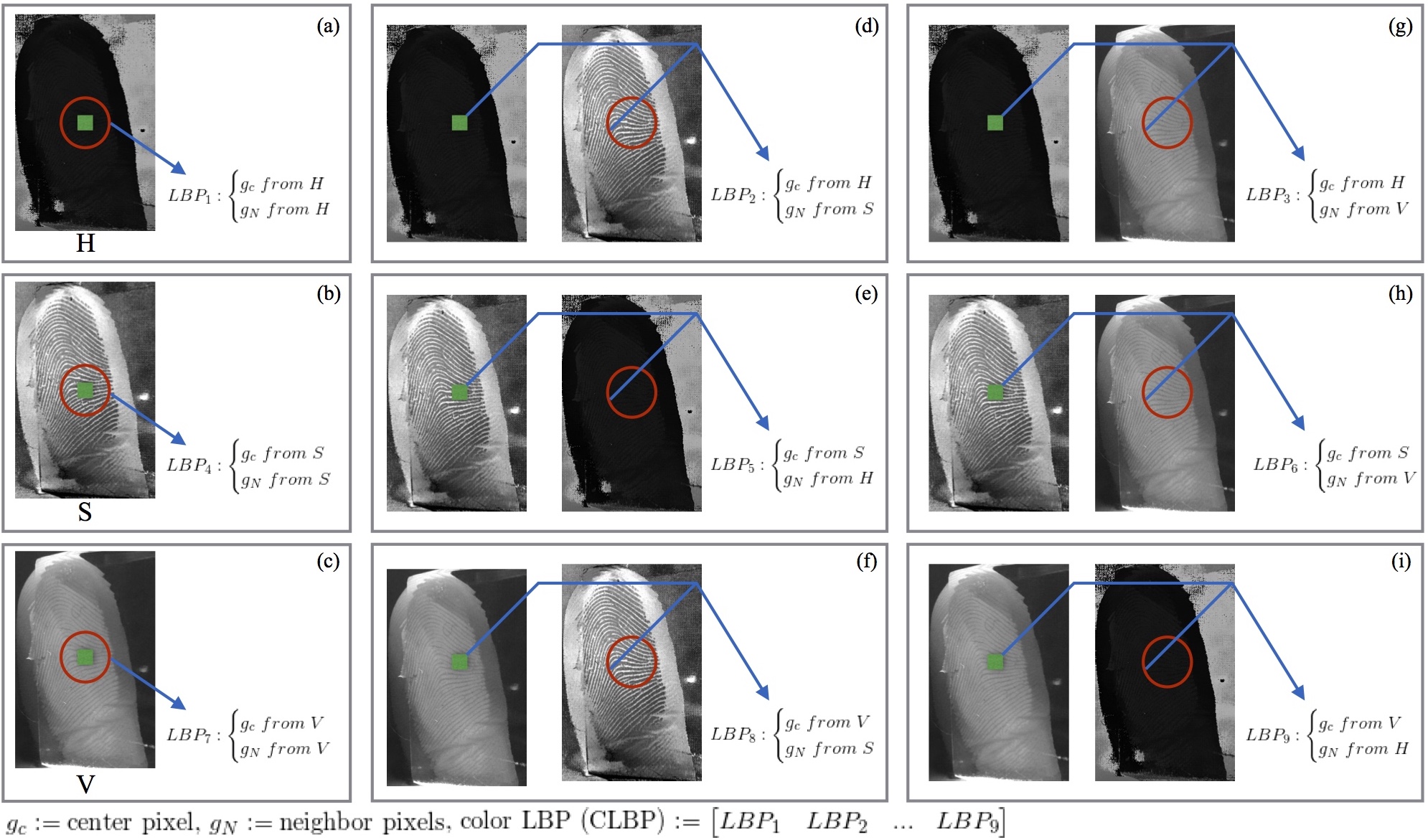}
\caption{Illustration of the extraction of the color LBP features from a spoof fingerprint impression captured by the RaspiReader. First, the fingerprint images from the RaspiReader are converted from the RGB color space to the HSV color space. Then, LBP as defined in equations 4, 5, and 6 is applied to (i) each individual channel of an image (a, b, c), and (ii) across opponent channels (e.g. a center pixel is taken from one channel and the neighborhood pixels are taken from another channel)(d, e, f, g, h, i). Finally, each of these LBP histograms are normalized and concatenated into a single feature vector. Note, the green squares illustrate the location of a center pixel and the red circles illustrate the location of neighboring pixels during each LBP extraction.}
\vspace{-1.5em}
\end{center}
\end{figure*}

Unlike traditional grayscale LBP patterns, color local binary patterns (CLBP) encode discriminative spatiochromatic textures from across multiple spectral channels \cite{color_lbp_rec}. In other words, CLBP extracts textures across all the different image bands (HSV used here) in a given input image. More formally, given an input image $I$ with $K$ spectral channels, let the set of all spectral channels for $I$ be defined as $S = \{S_1,...,S_K\}$. Then, the CLBP feature $\mathbf{x}$ can be extracted using algorithm 1. \begin{algorithm} 
\caption{Extraction of Color Local Binary Patterns} \begin{algorithmic}
   \STATE $\mathbf{x} \gets [\text{ }]$
    \FOR {$i\gets 1, K$}
    	\FOR {$j\gets 1, K$}
    		\STATE $\mathbf{x} \gets \mathbf{x} \Vert H_{8,1}(Si, Sj) \Vert H_{16,2}(Si, Sj) \Vert H_{24,3}(Si, Sj)$
	\ENDFOR
    \ENDFOR
    \RETURN $\mathbf{x}$
\end{algorithmic} \end{algorithm} Note that $H_{P,R}(g_c, g_N)$ returns a normalized histogram of local binary patterns using $g_c$ as the image channel that the center (thresholding) pixels are selected from, and $g_N$ as the image channel from which the neighborhood pixels are selected from in the computation of LBP as defined in equations 4, 5, and 6. Also, $\Vert$ indicates vector concatenation. This extraction of CLBP features is also visually presented in Figure 12. 

Using the live and spoof finger images from the RaspiReader's FTIR camera as described in Tables 2 and 3, five-fold cross validation is again used to partition training and testing splits. Then, we (i) convert each image to the HSV\footnote{Other color spaces were experimented with, but HSV consistently provided the highest performance. This is likely because HSV separates the luminance and chrominance components in an image, allowing extraction of LBP features on more complementary image channels (Fig. 14).} color space and (ii) extract 486 dimensional CLBP features from each image to train/test a binary linear SVM. With this proposed classification scheme and training/testing data, we report a mean (over five folds) True Detection Rate (TDR) of 90.75\% at a False Detection Rate (FDR) of 0.1\%. Additionally, we report the standard deviation for the TDR (over five folds) to be 8.88\%. 

To better understand which presentation attacks the spoof detector has a difficult time detecting using the proposed CLBP features and also how well the detector classifies live fingers, we report the mean (over five folds) correct detection rate for live fingers and spoofs (of every material) individually (Table 4).\begin{table}[h]
 \centering
 \begin{threeparttable}
\begin{tabular}{ |c||c|}
 \multicolumn{2}{c}{\specialcell{Table 4: Correct Detection Rates\tnote{1} on Live Fingers and Individual \\Spoof Materials Using RaspiReader Raw FTIR Images}} \\
 \hline
 Finger Type & \specialcell{Correct Detection Rate\\($\mu \pm \sigma$)} \\
 \hline
 \hline
 Live Finger & $95.07\% \pm 8.12$\\
 \hline
 \hline
 Ecoflex & $95.0\% \pm 5.0$ \\
 \hline
 \specialcell{Wood Glue} & $96.0\% \pm 6.52$\\
  \hline
 \specialcell{Monster Liquid Latex} & $100.0\% \pm 0.0$\\
 \hline
 \specialcell{Liquid Latex Body Paint} & $98.0\% \pm 4.47$\\
 \hline
 Gelatin & $100.0\% \pm 0.0$\\
 \hline
 \specialcell{Silver Coated Ecoflex} & $100.0\% \pm 0.0$\\
 \hline
 \specialcell{Crayola Model Magic} & $100.0\% \pm 0.0$\\
 \hline
\end{tabular}
\begin{tablenotes}
\item[1] The mean and standard deviation is reported for the correct detection rates over 5-folds, using 0.5 as a threshold (where scores range from 0-1).
\end{tablenotes}
\end{threeparttable}
%\vspace{-1.5em}
\end{table} From these results, we can see that that most difficult spoofs to detect using the CLBP features extracted from the RaspiReader raw FTIR images are ecoflex spoofs. This is to be expected since the ecoflex spoofs are quite thin. As such, much of the human skin color texture passes through the spoof material, making the spoof impression of similar color to live finger impressions (Fig. 15 (a)). We can also see from these results that the detector correctly classifies spoof fingers at a higher rate than live fingers. This is likely due to a small dataset of live fingers (only 15 human subjects provided live fingerprints) used during the training of the spoof detector. Given a larger number of live finger impressions to train on, the detection rates of live fingers will likely improve.     

Overall, in comparison to the baseline classifier (trained on grayscale LBP features extracted from processed images from a COTS fingerprint reader), this classifier (trained on CLBP features extracted from raw FTIR images from the RaspiReader) achieves significant improvement (90.75\% mean TDR compared to 58.33\% mean TDR both at FDR of 0.1\%). This huge jump in TDR from the baseline classifier at the same fixed FDR (0.1\%) demonstrates the usefulness of the raw FTIR images from the RaspiReader for fingerprint spoof detection. In particular, because the raw FTIR images from the RaspiReader contain color information, we are able to extract color textural (CLBP) features which are much more discriminative than traditional grayscale textural features.

However, a TDR of 90.75\% still leaves much room for improvement, especially if the proposed spoof detection scheme were to be installed in the field where a 9.25\% miss rate could prove very costly. Therefore, in the following experiment, we attempt to achieve even higher performance by using the direct image output of the RaspiReader for spoof detection.

\begin{figure}[t]
  \centering
  \subfloat[]{\includegraphics[scale=.225]{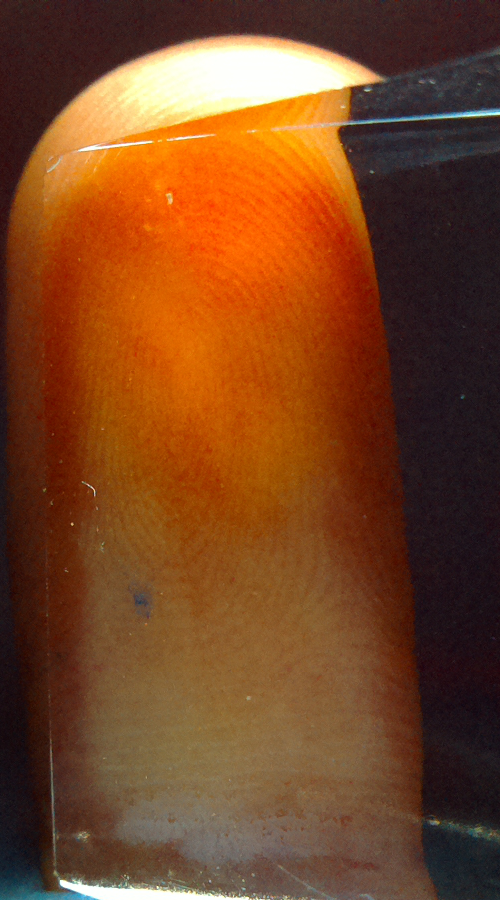}\label{fig:f1}}
  \hfill
  \subfloat[]{\includegraphics[scale=.225]{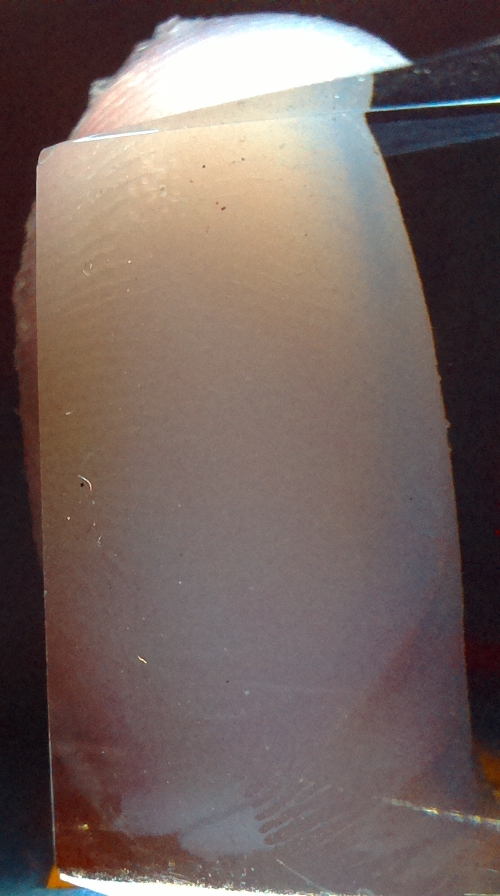}\label{fig:f2}}
  \caption{Raw direct fingerprint images acquired using the RaspiReader and used to conduct spoof detection Experiment 3. A live fingerprint impression is shown in (a). An ecoflex spoof fingerprint impression is shown in (b).}
\end{figure}

\begin{figure}[t]
  \centering
  \subfloat[]{\includegraphics[scale=.225]{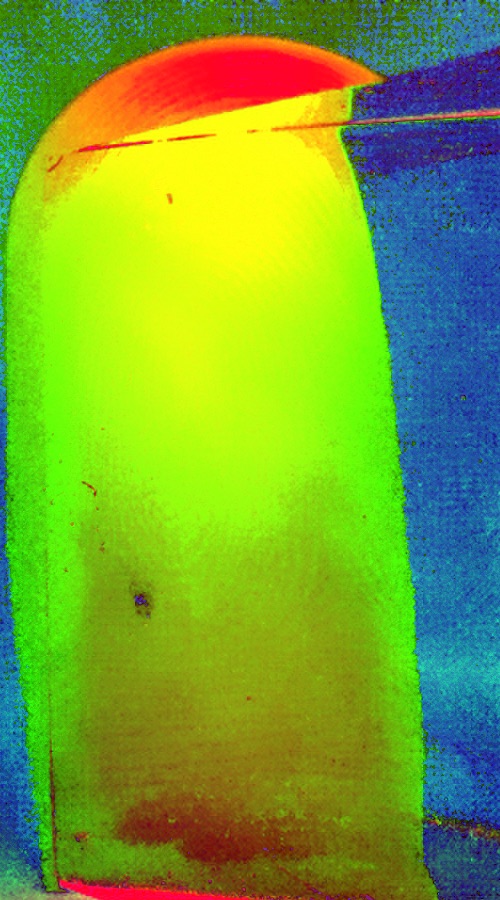}\label{fig:f1}}
  \hfill
  \subfloat[]{\includegraphics[scale=.225]{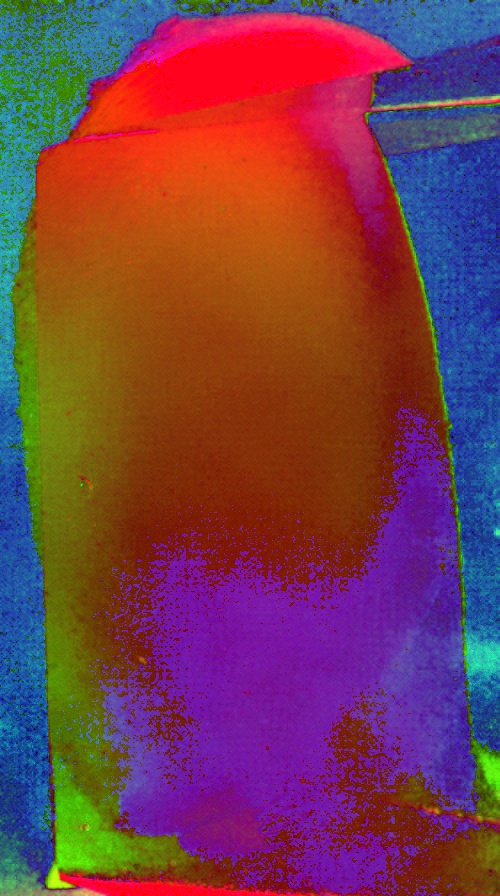}\label{fig:f2}}
  \caption{Direct fingerprint images acquired using the RaspiReader and converted to the HSV color space. A live fingerprint impression is shown in (a). An ecoflex spoof fingerprint impression is shown in (b).}
\end{figure}

{\it Experiment 3: Color LBP Extracted From RaspiReader Raw Direct Images.} To perform this experiment, we essentially repeat the entire structure of Experiment 2. However, instead of extracting CLBP features from the raw FTIR image output of the RaspiReader, 486 dimensional CLBP features are instead extracted from the direct image output of the RaspiReader (Fig. 13). 

Using the SVM classifier with the same parameters and 5-fold cross validation training/testing data, we report a mean (over five folds) TDR of 92.88\% at a FDR of 0.1\% (a slight improvement over the results of Experiment 2). Additionally, we report the standard deviation for the TDR to be 10.97\%. As in Experiment 2, we report the correct detection rates for live fingers and each individual spoof material separately to better determine how well the classifier correctly detects live fingers and each respective spoof type (Table 5).

\begin{table}[h]
 \centering
 \begin{threeparttable}
\begin{tabular}{ |c||c|}
 \multicolumn{2}{c}{\specialcell{Table 5: Correct Detection Rates\tnote{1} on Live Fingers and Individual \\Spoof Materials Using RaspiReader Raw Direct Images}} \\
 \hline
 Finger Type & \specialcell{Correct Detection Rate\\($\mu \pm \sigma$)} \\
 \hline
 \hline
 Live Finger & $98.8\% \pm 1.79$\\
 \hline
 \hline
 Ecoflex & $95.0\% \pm 8.66$ \\
 \hline
 \specialcell{Wood Glue} & $99.0\% \pm 2.24$\\
  \hline
 \specialcell{Monster Liquid Latex} & $100.0\% \pm 0.0$\\
 \hline
 \specialcell{Liquid Latex Body Paint} & $100.0\% \pm 0.0$\\
 \hline
 Gelatin & $99.0\% \pm 2.24$\\
 \hline
 \specialcell{Silver Coated Ecoflex} & $100.0\% \pm 0.0$\\
 \hline
 \specialcell{Crayola Model Magic} & $98.33\% \pm 3.73$\\
 \hline
\end{tabular}
\begin{tablenotes}
\item[1] The mean and standard deviation is reported for the correct detection rates over 5-folds, using 0.5 as a threshold (where scores range from 0-1).
\end{tablenotes}
\end{threeparttable}
\end{table}

\begin{figure}[t]
  \centering
  \subfloat[]{\includegraphics[scale=.225]{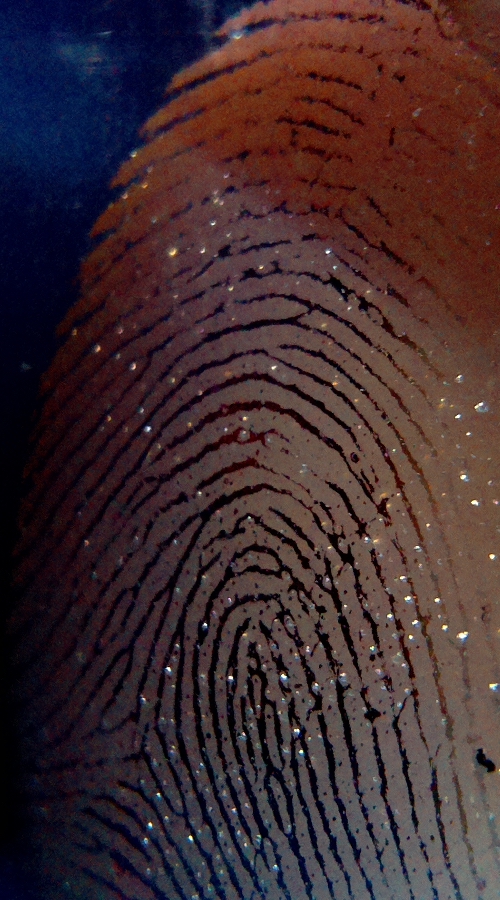}\label{fig:f1}}
  \hfill
  \subfloat[]{\includegraphics[scale=.225]{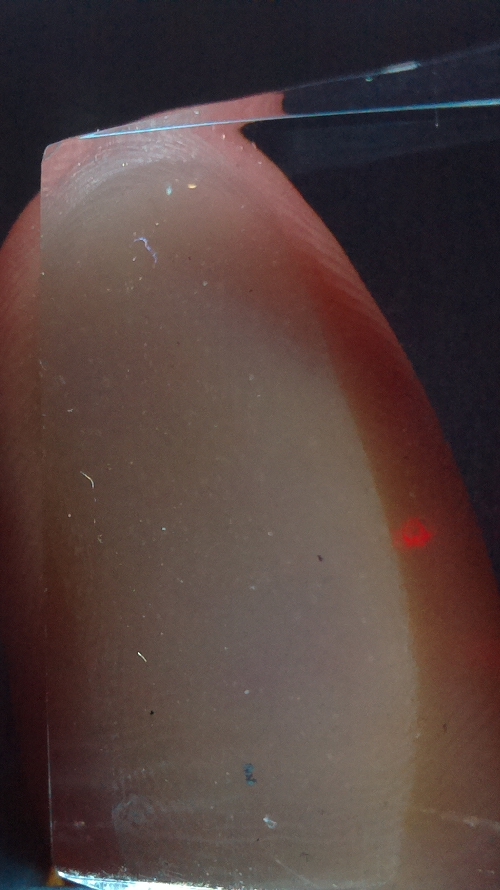}\label{fig:f2}}
  \caption{(a) Raw FTIR Image of an ecoflex spoof classified as live finger in Experiment 2. (b) Raw Direct Image of an ecoflex spoof classified as live finger in Experiment 3. Ecoflex is the most difficult spoof type for RaspiReader to defend against since human finger skin color passes through the clear, thin material. }
\end{figure}

From these results we can see that ecoflex spoofs are again the most difficult spoof type for RaspiReader to defend against since very thin clear spoofs allow much of the human skin color behind the spoof to pass through (Fig. 15 (b)). Also, in these direct images, if the surface area of the spoof is smaller than the surface area of the finger presenting the spoof (Fig. 15 (b)), much of the peripheral surface area of the live finger will be included in the spoof images used to make a classification decision, confusing the classifier. In the future, we plan to solve this problem by collecting more spoof impressions from thin clear spoofs, enabling us to better learn the more subtle differences between thin clear spoofs and live fingers. We can also see from these results that using the direct images of the RaspiReader enables higher correct detection rates of live fingers than the raw FTIR images do. This is likely because the raw direct images contain less variability amongst the various fingers than the raw FTIR images do. As such, the small amount of live finger training data does not affect performance as much in Experiment 3 as it does in Experiment 2.  

Finally, even at a TDR of 92.88\% in this experiment, there is still room for improvement. Therefore, in the final spoof detection experiment, we extract features from both of the RaspiReader images from camera-1 and camera-2  and concatenate them together in an attempt to achieve even higher and more robust spoof detection performance. 

{\it Experiment 4: Fusion of Color LBP Features.} In this penultimate spoof detection experiment, we extract CLBP features from both image outputs of the RaspiReader (raw FTIR output and raw direct output). These two CLBP features are then fused into a single 972 dimensional feature vector and used to train a binary linear SVM. 

Using this classification scheme, we report a mean (over five folds) TDR of 95.3\% at a FDR of 0.1\%. The standard deviation over the five folds was 6.05\%. By fusing CLBP features extracted from both the raw FTIR output of the RaspiReader and the direct output of the RaspiReader, we boost the classification performance beyond the performance when only a single image output from RaspiReader is used for spoof detection. This indicates that using two cameras for fingerprint image acquisition and subsequent spoof detection is advantageous, since they both contain complementary and discriminative spoof detection information. 

In summary, we have demonstrated in these spoof detection experiments that both raw images generated by our custom RaspiReader contain far more discriminative information for spoof detection than grayscale images output by COTS optical fingerprint readers. In addition, we have shown that these images contain complementary information such that when simple CLBP features extracted from each image are fused together, state-of-the-art spoof detection performance on known materials can be achieved. By showing that the RaspiReader images are far more useful for spoof detection than grayscale COTS optical images and by open sourcing the design of RaspiReader, we have removed what we posit to be one of the biggest limitations facing current state-of-the-art presentation attack detection schemes (namely low information processed images on which these detectors are trained). A full summary of these spoof detection experiment results is listed in Table 6.

\begin{table}[h]
 \centering
 \begin{threeparttable}
\begin{tabular}{ |c||c|c|}
 \multicolumn{3}{c}{Table 6: Summary of Spoof Detection Experiments} \\
 \hline
 Method & \specialcell{TDR @ FDR = 0.1\% \\ $\mu \pm \sigma$\tnote{1}} & \specialcell{Detection Time (msecs)\tnote{2}} \\
 \hline
 \hline
 \specialcell{$COTS_A$ \\+ LBP} & $58.33\% \pm 37.69$ & 236 \\
 \hline
 \specialcell{Rpi FTIR \\+ CLBP} & $90.75\% \pm 8.88$ & 243 \\
  \hline
 \specialcell{Rpi Direct \\+ CLBP} & $92.88\% \pm 10.97$ & 243 \\
 \hline
 \specialcell{Rpi Fusion \\+ CLBP} & $95.30\% \pm 6.05$ & 486 \\
 \hline
\end{tabular}
\begin{tablenotes}
\item[1] These results are reported over 5-folds.
\item[2] The LBP and CLBP feature extractions and SVM classifications were performed with a single CPU core on a Macbook Pro running a 2.9 GHz Intel Core i5 processor with 8 GB of RAM. In the future, we will parallelize the feature extractions and directly run the spoof detection algorithm on the Raspberry Pi controlling the RaspiReader.
\end{tablenotes}
\end{threeparttable}
\end{table}

\subsection{Interoperability of RaspiReader}

In addition to demonstrating the usefulness of the RaspiReader images for fingerprint spoof detection, we also demonstrate that by processing the RaspiReader FTIR images, we can output images which are compatible for matching with images from COTS fingerprint readers. Previously, we discussed how to process and transform a RaspiReader raw FTIR image into an image suitable for matching. In this experiment, we evaluate the matching performance (of 11,175 imposter pairs and 6750 genuine pairs) when using (i) the RaspiReader processed images as both the enrollment and probe images, (ii) the $COTS_A$ images as both the enrollment and probe images, and (iii) the $COTS_A$ images as the enrollment images and the RaspiReader processed images as the probe images. The results for these matching experiments are listed in Table 7. 

\begin{table}[h]
 \centering
\begin{tabular}{ |c|c|c|}
 \multicolumn{3}{c}{\specialcell{Table 7: Fingerprint Matching Results}} \\
 \hline
 Enrollment Reader & \specialcell{Probe Reader} &\specialcell{TAR @ FAR = 0.1\%} \\
 \hline
 \hline
 \specialcell{$COTS_A$} & $COTS_A$ & 98.62\% \\
 \hline
 \specialcell{RaspiReader} & RaspiReader & 99.21\% \\
  \hline
 \specialcell{$COTS_A$} & RaspiReader & 95.56\% \\
 \hline
\end{tabular}
\end{table}

From these results, we make two observations. First, the best performance is achieved for native comparisons, where the enrolled and search (probe) image are produced by the same capture device. RaspiReader's native performance is slightly better than that of $COTS_A$. This indicates that the RaspiReader is capable of outputting images which are compatible with state of the art fingerprint matchers\footnote{We use the Innovatrics fingerprint SDK since we recently acquired this matcher, and it is shown to have high accuracy in the NIST FpVTE evaluation \cite{nist}.}. Second, we note that the performance does drop slightly when conducting the interoperability experiment ($COTS_A$ is used for enrollment images and RaspiReader is used for probe images). However, the matching performance is still quite high considering the very stringent operating point (FAR = 0.1\%). Furthermore, studies have shown that when different fingerprint readers are used for enrollment and subsequent verification or identification, the matching performance indeed drops \cite{ross_cross,cross,engelsma2017universal}. Finally, we are currently investigating other approaches for processing and downsampling RaspiReader images to reduce some of the drop in cross-reader performance. 

\section{Conclusions}
We have designed and prototyped a custom fingerprint reader, called RaspiReader, with ubiquitous components. This fingerprint reader is both low cost and easy to assemble, enabling other researchers to easily and seamlessly develop their own novel fingerprint presentation attack detection solutions which use both hardware and software. By customizing RaspiReader with two cameras for fingerprint image acquisition rather than one, we were able to extract discriminative color local binary patterns (CLBP) from both raw images which, when fused together, enabled us to match the performance of state of the art spoof detection methods (CNN). Finally, by processing the raw FTIR images of the RaspiReader, we were able to output fingerprint images compatible for matching with COTS optical fingerprint readers. 

In our ongoing work, we plan to integrate specialized hardware into RaspiReader such as Optical Coherence Tomography (OCT) for sub-dermal imagery, IR cameras for vein detection, or microscopes for extremely high resolution images of the fingerprint. Because the RaspiReader uses ubiquitous components running open source software, RaspiReader enables future integration of these additional hardware components. In addition to the integration of specialized hardware, we also plan to use the raw, information rich images from the RaspiReader to pursue one-class classification schemes for fingerprint spoof detection. In particular, we posit that the RaspiReader images will assist us in modeling the class of live fingerprint images, such that spoofs of all material types can be easily rejected. 

% or


\section*{Acknowledgment}
This research was supported by both grant no. 70NANB17H027 from the NIST Measurement Science program and also by the Office of the Director of National Intelligence (ODNI), Intelligence Advanced Research Projects Activity (IARPA), via IARPA R\&D Contract No. 2017 - 17020200004. The
views and conclusions contained herein are those of the
authors and should not be interpreted as necessarily representing
the official policies, either expressed or implied,
of ODNI, IARPA, or the U.S. Government. The U.S. Government
is authorized to reproduce and distribute reprints
for governmental purposes notwithstanding any copyright
annotation therein.

\ifCLASSOPTIONcaptionsoff
  \newpage
\fi



%\bibliographystyle{IEEEtran}
\bibliography{submission_example}
\bibliographystyle{ieeetr}


\vspace{-10.0 mm}
%\vfill
\begin{IEEEbiography}[{\includegraphics[width=1in,height=1.25in,clip,keepaspectratio]{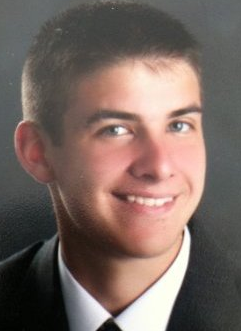}}]{Joshua J. Engelsma}
graduated magna cum laude with a B.S. degree
in computer science from Grand Valley State University, 
Allendale, Michigan, in 2016. He is currently
working towards a PhD degree in the
Department of Computer Science and Engineering
at Michigan State University, East Lansing,
Michigan. His research interests include pattern
recognition, computer vision, and image processing
with applications in biometrics.
\end{IEEEbiography}

%\vspace{-10.0 mm}
\begin{IEEEbiography}[{\includegraphics[width=1in,height=1.25in,clip,keepaspectratio]{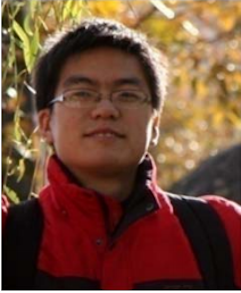}}]{Kai Cao}
received the Ph.D. degree from the
Key Laboratory of Complex Systems and Intelligence
Science, Institute of Automation, Chinese
Academy of Sciences, Beijing, China, in 2010.
He is currently a Post Doctoral Fellow in the Department
of Computer Science \& Engineering,
Michigan State University. He was affiliated with
Xidian University as an Associate Professor. His
research interests include biometric recognition,
image processing and machine learning.
\end{IEEEbiography}

%\newpage
%\vspace{-20.0 mm}
\begin{IEEEbiography}[{\includegraphics[width=1in,height=1.25in,clip,keepaspectratio]{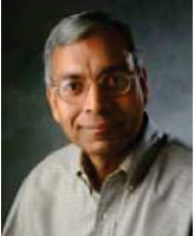}}]{Anil K. Jain}
is a University distinguished professor
in the Department of Computer Science
and Engineering at Michigan State University.
His research interests include pattern recognition
and biometric authentication. He served as
the editor-in-chief of the IEEE Transactions on
Pattern Analysis and Machine Intelligence and
was a member of the United States Defense Science
Board. He has received Fulbright, Guggenheim,
Alexander von Humboldt, and IAPR King
Sun Fu awards. He is a member of the National
Academy of Engineering and foreign fellow of the Indian National
Academy of Engineering.
\end{IEEEbiography}
\vfill


%\vfill


\end{document}